\documentclass[
twocolumn,
]{ceurart}

\usepackage{listings}
\usepackage{algorithm}
\usepackage{algorithmic}
\usepackage{booktabs}
\usepackage{tabularx}   
\usepackage{array}  
\usepackage{newfloat}
\usepackage{listings}
\usepackage{tcolorbox}
\usepackage{subcaption}

\makeatletter
\renewcommand{\xmpdatawrite}[2]{}
\makeatother

\begin{document}


\conference{}

\title{Personality Expression Across Contexts: Linguistic and Behavioral Variation in LLM Agents}

\author{Bin Han}
\author{Deuksin Kwon}
\author{Jonathan Gratch}
\address{University of Southern California}


\begin{abstract}
Large Language Models (LLMs) can be conditioned with explicit personality prompts, yet their behavioral realization often varies depending on context.
This study examines how identical personality prompts lead to distinct linguistic, behavioral, and emotional outcomes across four conversational settings—ice-breaking, negotiation, group decision, and empathy tasks.
Results show that contextual cues systematically influence both personality expression and emotional tone, suggesting that the same traits are expressed differently depending on social and affective demands.
This raises an important question for LLM-based dialogue agents: whether such variations reflect inconsistency or context-sensitive adaptation akin to human behavior.
Viewed through the lens of Whole Trait Theory, these findings highlight that LLMs exhibit context-sensitive rather than fixed personality expression, adapting flexibly to social interaction goals and affective conditions.
\end{abstract}


\maketitle

\section{Introduction}


Large language models (LLMs) have recently been shown to support increasingly complex forms of social interaction, including reasoning about context, emotion, and strategic behavior~\cite{thapa2025large, lee2025since, tak2023gpt,kwon2024llms}. Recent progress in LLMs has enabled conversational agents to exhibit distinct personality characteristics during interaction~\cite{jiang2023evaluating,serapio2023personality}.
Beyond improving linguistic performance, research has increasingly focused on enhancing the social quality of communication, aiming to make agents appear more human-like and engaging.
Several studies have demonstrated that personality-conditioned agents can enhance user trust, engagement, and conversational satisfaction~\cite{lee2006can,ait2023power}.
For example, Ait Baha et al.~\cite{ait2023power} conducted a systematic review showing that personality-adaptive chatbots significantly improve user satisfaction and engagement.
More recent work moves beyond surface-level imitation, showing that LLMs can coherently understand and reproduce personality constructs.
For instance, Extraverted agents tend to use more positive emotion and social words, while conscientious agents favor structured and goal-oriented expressions, demonstrating linguistic and emotional consistency with their assigned traits~\cite{banayeeanzade2025psychological, jiang2023personallm}.
These influences extend beyond linguistic expression to shape decision-making styles and even nonverbal expressivity~\cite{huang2024personality, hale2025kodis, han2025can}.

However, emerging findings suggest that personality expression in LLMs is not always stable across contexts.
Even under identical persona/personality prompts, personality can be expressed quite differently depending on the context~\cite{ han2025can,reusens2025economists}.
For example, an agent instructed to be extroverted may display humor and expressiveness in small talk, but adopt a more neutral and goal-oriented style in negotiation~\cite{han2025can}. 
Reusens et al.~\cite{reusens2025economists} similarly showed that personality-aligned LLMs exhibit varying personality expression across task contexts, with consistency increasing in more structured settings but decreasing in open-ended conversations.
This raises an important issue for LLM-based dialogue agents: are such variations best understood as a lack of consistency, or as a natural consequence of context-sensitive modulation similar to human behavior?

Personality traits, such as those captured by the Big Five, represent people’s average tendencies that are expressed across a variety of situations~\cite{john1999big}. 
However, psychological research shows that a person’s actual behavior can still shift depending on the situation, goals, and social context~\cite{mischel1995cognitive}.
\textbf{\textit{Whole Trait Theory}} integrates both perspectives by conceptualizing personality as a distribution of momentary states—reflecting average stability—whose variability arises from underlying social-cognitive mechanisms such as goals, beliefs, and affect~\cite{fleeson2001toward,fleeson2015whole}.
This perspective suggests that the variability in personality expression observed in LLMs may not indicate inconsistency, but rather could be seen as reflecting the contextual flexibility of human personality.
Similar computational frameworks have conceptualized personality as emergent from dynamic motivational and neural systems, providing theoretical grounding for such context-sensitive adaptation~\cite{read2021neural,read2025virtual}.

To better understand this phenomenon, we conduct a systematic analysis of context-dependent personality expression in LLM-based dialogue agents.
Our study covers four dialogue contexts—small talk, negotiation, survival planning, and empathetic dialogue—which differ in social goal, level of cooperation, and emotional tone.
We analyze how personality expression emerges through linguistic style, behavioral patterns, and task outcomes. This allows us to examine not only whether personality expression changes across contexts, but also whether such changes show consistent alignment between linguistic and behavioral dimensions.

\begin{figure}
    \centering
    \includegraphics[width=1\linewidth]{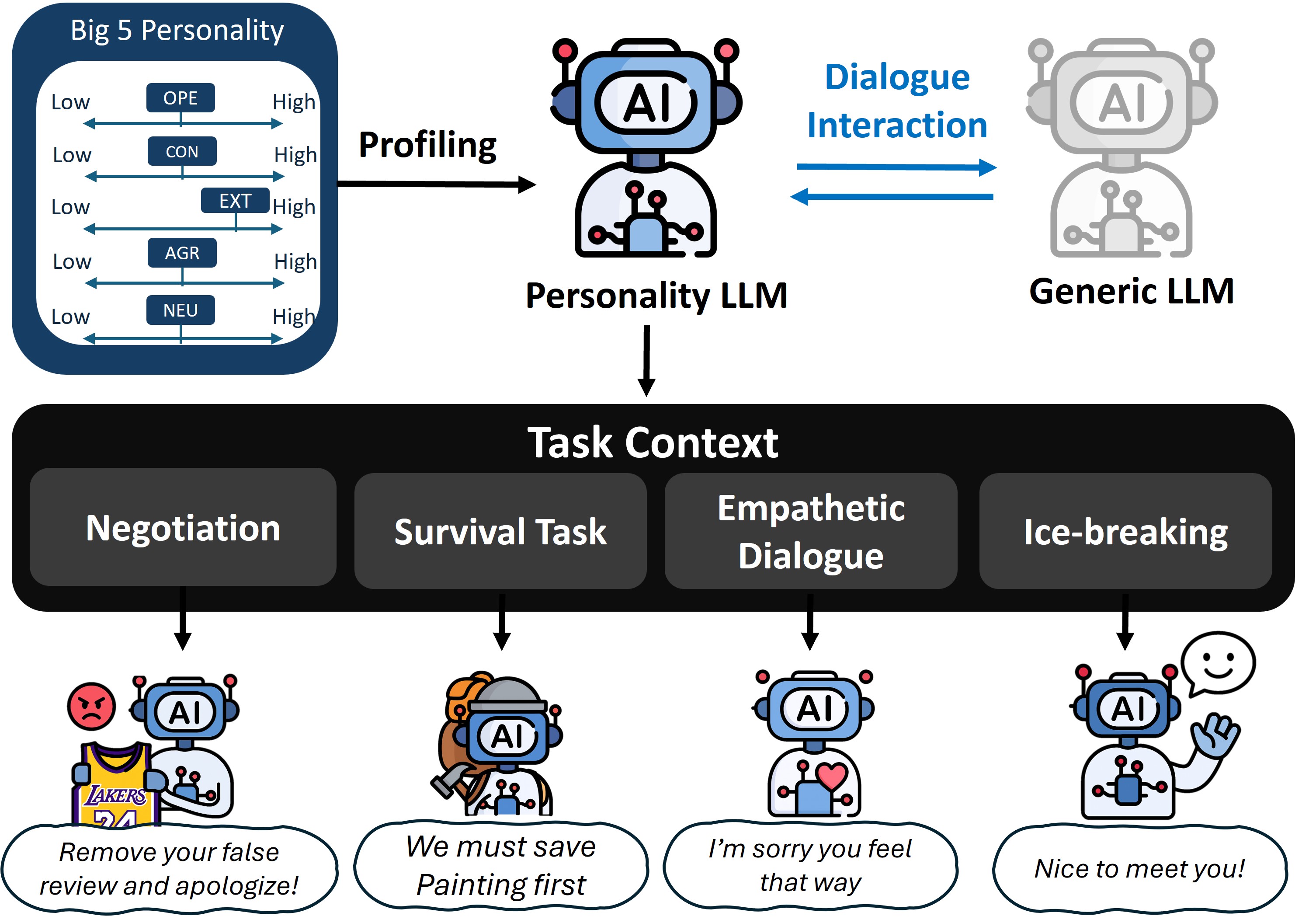}
    \caption{Overview of the experimental framework examining context-dependent personality expression in LLM-based dialogue agents}
    \label{fig:overview}
\end{figure}

\section{Research Questions}

\begin{itemize}
\item \textbf{RQ1. Linguistic expression of personality:}  
How does personality expression differ across dialogue contexts (ice-breaking, negotiation, survival, empathy), 
and which linguistic and emotional cues account for these contextual variations?

\item \textbf{RQ2. Behavioral expression of personality:}  
Do personality-driven tendencies extend beyond linguistic style to task-oriented behaviors such as concession-making, and agreement outcomes?
\end{itemize}

\section{Related Work}

\subsection{Personality and LLMs}

Recent research increasingly explores how LLMs can be guided to exhibit specific personality traits.
Personality modulation in LLMs has been achieved through various approaches, including personality prompting and instruction tuning~\cite{jiang2023personallm,serapio2023personality}.
Prior work also shows that LLMs can emulate human-like traits such as trust, personality, or emotion-driven behavior \cite{kwon2025evaluating}.
Furthermore, researchers have begun to apply standardized psychometric instruments—originally developed for human assessment, such as the BFI~\cite{john1999big} and IPIP-NEO~\cite{goldberg1999broad}—to examine how LLMs interpret and express personality-related constructs~\cite{jiang2023evaluating}.
Despite these advances, most existing evaluations focus on single-turn offering limited insight into how personality is expressed across dialogue contexts.

\subsection{Context-Dependent Personality Expression in Humans}

In contrast, personality psychology has long documented that while traits predict stable average tendencies, the actual expression of those traits is highly context-dependent. 
For example, extraverted individuals may show heightened sociability in affiliation-oriented settings, but display increased assertiveness or even aggression in competitive situations such as negotiations.
Conscientiousness tends to increase under achievement goals, while agreeableness and dominance adapt flexibly to a partner’s interpersonal style \cite{kiesler1996contemporary,moskowitz1994cross,mccabe2016traits}.
Whole Trait Theory formalizes this duality, treating traits as density distributions of states (stable on average) whose variability is systematically shaped by goals, situational cues, and interaction partners \cite{fleeson2001toward,fleeson2015whole}. 
Related frameworks such as trait activation theory \cite{tett2003personality} and interpersonal theory \cite{kiesler1996contemporary} similarly emphasize that traits are not rigid prescriptions but potentials expressed when contexts activate trait-relevant goals or social schemas. 
This literature suggests that personality expression is best understood as context-sensitive rather than fixed. 
Yet LLM research has not fully incorporated this perspective, often evaluating persona conditioning only in static or decontextualized settings. Our work seeks to bridge this gap by drawing on psychological theory and systematically testing how context shapes personality expression in LLM-based dialogue agents.

\section{Experimental Design}

\subsection{Personality Conditions}

We manipulate LLM personality expression using personality prompts adopted from prior studies on personality-conditioned dialogue agents~\cite{serapio2023personality,kwon2025evaluating}.
Each prompt explicitly defines a single Big Five trait at two levels — \textit{High} and \textit{Low} — using wording adapted from these prior works~\cite{serapio2023personality,jiang2023personallm,noh2024llms}.
The phrasing follows a descriptive format such as:
\textit{“You are a highly extroverted person: energetic, sociable, talkative, and enthusiastic”}
or
\textit{“You are an introverted person: reserved, quiet, and reflective”}.
The adjectives representing each trait’s major facets were also drawn from established personality lexicons~\cite{, jiang2023personallm,costa1995domains},
ensuring consistency with prior personality-prompting studies.
This manipulation has been validated in previous work using the BFI-10~\cite{john1999big}, confirming that such textual prompts reliably elicit consistent personality representations in LLMs.

\subsection{Agent Simulation: Personality vs. Generic Agent}

All experiments were conducted in an agent–agent interaction environment.
One agent was personality-driven according to the designated persona prompt (referred to as the Personality Agent), while the other served as a Generic Agent providing a neutral baseline for comparison (without personality prompting).
Because the partner’s personality can also influence interaction dynamics, we kept it constant across all experiments by using a neutral (non-personalized) agent.
This configuration follows the frameworks~\cite{han2025can, kwon2025evaluating}.

\subsection{Main Tasks}
We evaluate personality expression across four dialogue contexts, each chosen to highlight different situational properties of interaction (see Table~\ref{tab:dialogue_contexts}):

\begin{table*}[h]
\centering
\caption{Comparison of dialogue contexts by dominant emotion, social goal~\cite{mcclelland1987human}, and cooperation level.}
\begin{tabular}{p{3.0cm}p{4.5cm}p{5.2cm}p{3.0cm}}
\toprule
\textbf{Task} & \textbf{Emotion} & \textbf{Dominant Social Goal~\cite{mcclelland1987human}} & \textbf{Cooperation Level} \\
\midrule
Ice-breaking & Joy (Valence $\uparrow$, Arousal $\uparrow$) & \textbf{Affiliation} – rapport building and self-disclosure & Medium–High \\
Negotiation & Anger (Valence $\downarrow$, Arousal $\uparrow$) & \textbf{Power} – assertion and influence & Low–Medium \\
Survival Task & Neutral (Valence $-$, Arousal $-$) & \textbf{Achievement} – joint problem solving and coordination & High \\
Empathetic Dialogue & Sadness (Valence $\downarrow$, Arousal $\downarrow$) & \textbf{Affiliation} – emotional support and care & Very High \\
\bottomrule
\end{tabular}
\label{tab:dialogue_contexts}
\end{table*}

\begin{itemize}
    \item \textbf{Task 1: Ice-breaking}  
    This task is designed to elicit friendly, casual, and emotionally positive interactions between agents.  
    We adopted the question set from the \textit{Personal Questions} paradigm~\cite{lee2006can, aron1997experimental}, which has been widely used in prior human--robot interaction research.  
    In this task, the Personality Agent responds to three casual questions from the Generic Agent (e.g., ``What do you like to do in your free time?'').  
    Rather than focusing on solving a shared problem, this task centers on sustaining natural conversation and promoting self-disclosure.  
    The overall atmosphere is warm, informal, and cooperative, encouraging open and engaging dialogue.

    \item \textbf{Task 2: Negotiation}  
    This task is based on the KODIS dataset~\cite{hale2025kodis}, which models buyer--seller disputes in resource-allocation scenarios.  
    The Personality Agent takes the role of a buyer requesting a refund, while the Generic Agent plays the seller seeking the removal of a negative review.  
    Following prior work~\cite{hale2025kodis}, we assigned the buyer role to the Personality Agent, as this position tends to exhibit greater emotional dynamics and expressive variability during negotiation.  
    The interaction involves multiple issues---refund amount, review removal, and apology---making it suitable for observing complex behavioral strategies and adaptive negotiation patterns.  
    The emotional tone is dominated by anger and frustration, characterized by high tension, low cooperation, and a formal, goal-oriented dialogue style.

    \item \textbf{Task 3: Survival Task (Group Decision Making)}  
    This task is adapted from Artstein et al.~\cite{artstein2017listen}, originally involving 15 artworks to be rescued from a museum fire.  
    To focus more clearly on consensus-building behaviors, we reduced the number of items to five.  
    Each agent begins with opposing initial rankings (A, B, C, D, E for the Personality Agent and E, D, C, B, A for the Generic Agent), and they must justify their choices and negotiate to reach an agreement.
    This task emphasizes a collaborative decision-making setting in which both agents act as a team to achieve a shared goal.  
    The overall tone is calm and constructive, with moderate formality, high cooperation, and a positive emotional.

    \item \textbf{Task 4: Empathetic Dialogue}  
    This task is based on the \textit{Empathetic Dialogues} dataset~\cite{rashkin2018towards}, designed to evaluate the agent's ability to generate personality-consistent empathetic responses.  
    The Generic Agent presents an emotionally charged statement (e.g., ``I've been feeling so lost since I failed the certification exam again.''), and the Personality Agent must respond with emotional understanding and supportive expression.  
    The conversation context is emotionally sensitive and personal, characterized by high affective engagement, low formality, and a supportive atmosphere.
\end{itemize}

\subsection{Evaluation}

\subsubsection{Linguistic Measures}
To capture language-level cues associated with personality expression, we employ:

\begin{itemize}
\item \textbf{Linguistic Inquiry and Word Count (LIWC):}
We use the widely adopted psycholinguistic text analysis toolkit, Linguistic Inquiry and Word Count (LIWC)~\cite{pennebaker2015development}, to examine lexical and stylistic differences between personality conditions. LIWC provides a context-free linguistic analysis, as it captures lexical and stylistic patterns based solely on word usage without incorporating dialogue context.

\item \textbf{Personality Prediction (Pre-trained Classifier):}  
To assess trait expression quantitatively, we employ a pre-trained Big Five personality classifier~\cite{kazameini2020personality} that integrates contextualized embeddings from BERT with psycholinguistic features.  
The model predicts the likelihood that a given utterance reflects a specific trait (e.g., Extraversion = 1 if classified as extroverted, 0 otherwise). Similarly, the classifier offers a context-free estimate of personality based only on linguistic feature.

\item \textbf{LLM-based Personality Evaluation:}  
We further examined perceived personality by prompting an LLM to act as an expert personality psychologist and evaluate each conversation on the Big Five traits (1–5 scale) based on its dialogue context~\cite{jiang2023personallm}.
The evaluation prompt included explicit criteria for each trait and score level, outlining behavioral expectations for high versus low expressions.  
The task context was explicitly provided in the prompt to enable context-dependent evaluation.  
Following a third-person annotation approach~\cite{huang2025beyond}, the model produced both numerical ratings and concise rationales for each dialogue.
\end{itemize}

\subsubsection{Emotion Measures}
We analyze the emotion that emerged during interactions. 
While LIWC captures affective word usage at the lexical level, 
it does not fully reflect the overall emotional tone or intensity conveyed across dialogue turns. 
To complement such surface-level measures, we explicitly evaluate the affective tone expressed by each agent using an LLM-based emotion recognition method~\cite{kwon2025evaluating,han2024knowledge,tak2024gpt}. 
Given that recent studies have demonstrated the strong capability of LLMs in emotion reasoning and affective understanding,
we leveraged this approach to perform a comprehensive assessment of each agent’s expressed affect. 
The model inferred the overall affective state of the speaker by integrating the linguistic content and stylistic tone of utterances, 
and the output was represented in complementary affective dimensions (Valence and Arousal~\cite{russell1980circumplex}).

\subsubsection{Behavioral (Decision) Measures}

In task-oriented contexts such as \textbf{Negotiation} and \textbf{Survival (Save the Art)},  
we analyzed two behavioral indicators of cooperation:  
(1) whether the two agents ultimately reached a mutual agreement (\textit{Agreement rate}), and  
(2) how much each agent adjusted its decision in response to the partner’s behavior (\textit{Concession}).  

\begin{itemize}
    \item \textbf{Negotiation (Refund Offer)}:  
        In the negotiation task, concession-making was defined as the reduction from the initial refund proposal,  
        quantified as 
        \[
        \text{Concession} = 100\% - \text{Refund Offer}.
        \]
        This measure captures how much an agent moved from its original 100\% offer toward the partner’s demand across dialogue rounds.  
        Plotting this value over turns yields a \textit{concession curve}, which represents the trajectory of compromise throughout the negotiation.

    \item \textbf{Survival (Sum of Rank Differences; SRD):}
    In the survival task, concession-making at round \(t\) was measured as the
    difference between the current ranking and the initial ranking.
    Specifically, we computed the \textit{Sum of Rank Differences (SRD)}~\cite{heberger2010sum}
    for each round \(t\) as
    \[
    \text{SRD}_{t} = \sum_{i=1}^{5} \big| r_{i}^{\text{initial}} - r_{i}^{(t)} \big|,
    \]
    where \(r_{i}^{\text{initial}}\) and \(r_{i}^{(t)}\) denote the ranks offered by the personality agent for item \(i\)
    at the initial and current rounds, respectively.
    Because all agents begin with the same \textit{initial order}, the SRD value starts at \(0\)
    and increases as the ranking deviates from the baseline.
    A higher \(\text{SRD}_{t}\) therefore indicates a larger deviation from the initial decision state
    at that round (i.e., greater concession relative to the baseline).
\end{itemize}

\section{Result: Personality Across Contexts}

We analyze personality expression by comparing High and Low personality conditions across the dialogue contexts. 
The following subsections present an analysis of how personality expression interacts with dialogue context.

\subsection{Linguistic Inquiry and Word Count (LIWC)}

Among the many LIWC features, we selected only those shown to significantly correlate with the Big Five traits in prior meta-analysis~\cite{koutsoumpis2022kernel}.
Table~\ref{tab:liwc_task_comparison} summarizes the mean differences between High and Low groups across tasks.
The observed directional trends all followed patterns reported in previous findings~\cite{koutsoumpis2022kernel}.

\renewcommand{\arraystretch}{0.8}
\begin{table*}[h]
\centering
\scriptsize
\begin{tabularx}{\textwidth}{l l *{8}{>{\centering\arraybackslash}X} >{\centering\arraybackslash}p{2.6cm}}
\toprule
\textbf{Trait} & \textbf{LIWC Feature} &
\multicolumn{2}{c}{\textbf{Task 1}} &
\multicolumn{2}{c}{\textbf{Task 2}} &
\multicolumn{2}{c}{\textbf{Task 3}} &
\multicolumn{2}{c}{\textbf{Task 4}} &
\textbf{Reported} \\
\cmidrule(lr){3-4} \cmidrule(lr){5-6} \cmidrule(lr){7-8} \cmidrule(lr){9-10}
 &  & High & Low & High & Low & High & Low & High & Low &
 {\scriptsize \raisebox{-0.6ex}{(\cite{koutsoumpis2022kernel})}} \\[+0.6ex]
\toprule
Openness & Word Count/sentence & \textbf{18.0} & 14.0 & \textbf{15.7} & 14.5 & \textbf{16.3} & 14.3 & \textbf{17.2} & 14.7 & High $>$ Low \\
 & Leisure & \textbf{1.2} & 1.0 & \textbf{1.1} & 1.0 & \textbf{1.1} & 1.0 & \textbf{1.1} & 1.0 & High $>$ Low \\
\cmidrule(lr){1-11}
Conscientiousness & Swear & 0.0 & \textbf{1.0} & 0.0 & \textbf{1.0} & 0.0 & \textbf{1.0} & 0.0 & \textbf{1.0} & High $<$ Low \\
 & Anger & 1.0 & \textbf{1.3} & 1.0 & \textbf{1.2} & 1.1 & \textbf{1.2} & 1.0 & \textbf{1.2} & High $<$ Low \\
 & Negative Emotions & 2.9 & \textbf{3.9} & 3.0 & \textbf{3.8} & 3.0 & \textbf{3.6} & 2.9 & \textbf{3.6} & High $<$ Low \\
 & Biological Processes & 1.2 & \textbf{1.3} & 1.1 & \textbf{1.3} & 1.2 & \textbf{1.3} & 1.2 & \textbf{1.3} & High $<$ Low \\
 & Pronouns & 18.2 & \textbf{20.8} & 18.0 & \textbf{20.5} & 18.0 & \textbf{20.5} & 18.0 & \textbf{20.3} & High $<$ Low \\
 & Cognitive Processes & \textbf{16.1} & 14.8 & \textbf{15.8} & 13.9 & \textbf{15.9} & 14.0 & \textbf{15.8} & 14.0 & High $>$ Low \\
 & Tentative & 1.3 & \textbf{1.8} & 1.3 & \textbf{1.8} & 1.3 & \textbf{1.7} & 1.3 & \textbf{1.7} & High $<$ Low \\
\cmidrule(lr){1-11}
Extraversion & Word Count/sentence & \textbf{17.6} & 14.2 & \textbf{15.5} & 14.0 & \textbf{16.0} & 14.0 & \textbf{16.8} & 14.5 & High $>$ Low \\
 & Sexual & \textbf{1.0} & 0.9 & \textbf{1.0} & 0.9 & \textbf{1.0} & 0.9 & \textbf{1.0} & 0.9 & High $>$ Low \\
\cmidrule(lr){1-11}
Agreeableness & Swear & 0.0 & \textbf{1.0} & 0.0 & \textbf{1.0} & 0.0 & \textbf{1.0} & 0.0 & \textbf{1.0} & High $<$ Low \\
 & Anger & 1.0 & \textbf{1.3} & 1.0 & \textbf{1.3} & 1.0 & \textbf{1.3} & 1.0 & \textbf{1.2} & High $<$ Low \\
 & Positive Emotions & \textbf{10.3} & 4.6 & \textbf{10.3} & 4.6 & \textbf{10.4} & 4.7 & \textbf{10.3} & 4.7 & High $>$ Low \\
\cmidrule(lr){1-11}
Neuroticism & Negative Emotions & 2.5 & \textbf{4.3} & 2.5 & \textbf{4.1} & 2.5 & \textbf{4.2} & 2.5 & \textbf{4.1} & High $<$ Low \\
 & I & 5.2 & \textbf{9.2} & 5.2 & \textbf{8.8} & 5.1 & \textbf{8.7} & 5.1 & \textbf{8.6} & High $<$ Low \\
 & Pronouns & 16.0 & \textbf{23.1} & 16.9 & \textbf{22.7} & 16.3 & \textbf{22.3} & 16.1 & \textbf{22.2} & High $<$ Low \\
\bottomrule
\end{tabularx}
\vspace{0.3em}
\caption{Mean LIWC feature values for High vs. Low personality across the four dialogue contexts (Task 1: Ice-breaking, 2: Negotiation, 3: Survival, 4: Empathetic Conversation). Bold numbers represent the higher mean within each pair, showing directional trends consistent with prior findings \cite{koutsoumpis2022kernel}.}
\label{tab:liwc_task_comparison}
\end{table*}

\subsection{Pre-trained Personality Prediction}

\begin{figure*}[h]
    \centering
    \includegraphics[width=1\linewidth]{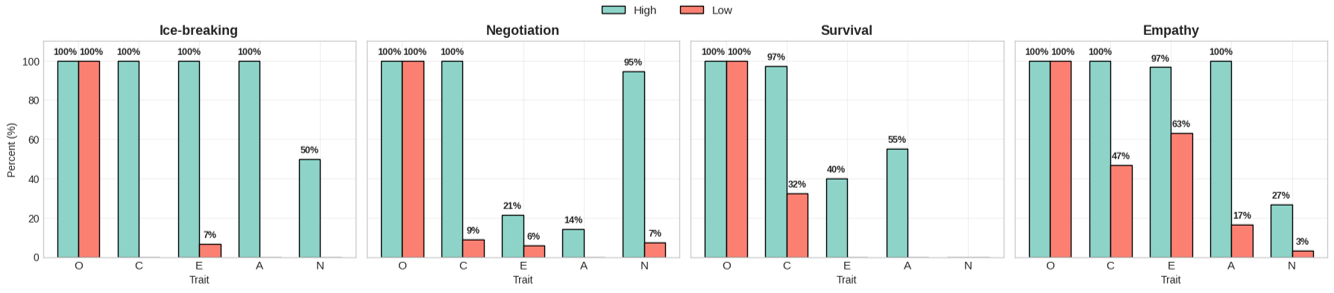}
    \caption{BERT-based Pre-trained Personality Prediction Model Result~\cite{kazameini2020personality}}
    \label{fig:pre-train}
\end{figure*}

Figure~\ref{fig:pre-train} shows the proportion of utterances predicted as expressing each Big Five trait (High vs. Low) by the pre-trained personality classifier~\cite{kazameini2020personality} across four dialogue contexts. 
The overall direction of prediction aligned well with the intended personality prompts, indicating that the model captured the trait differences embedded in the agents’ language. 
While the general trend showed agreement with the designed manipulation, some divergence was also observed—reflecting the challenge of relying on context-ignorant linguistic measures. 
Trait separation was most evident in the Ice-breaking task, where expressive and affiliative language facilitated clearer distinctions, whereas Negotiation and Survival showed broadly similar patterns with reduced differentiation due to their more goal-directed nature.

\subsection{LLM-based Personality Evaluation}

\begin{figure*}[h]
    \centering
    \includegraphics[width=1\linewidth]{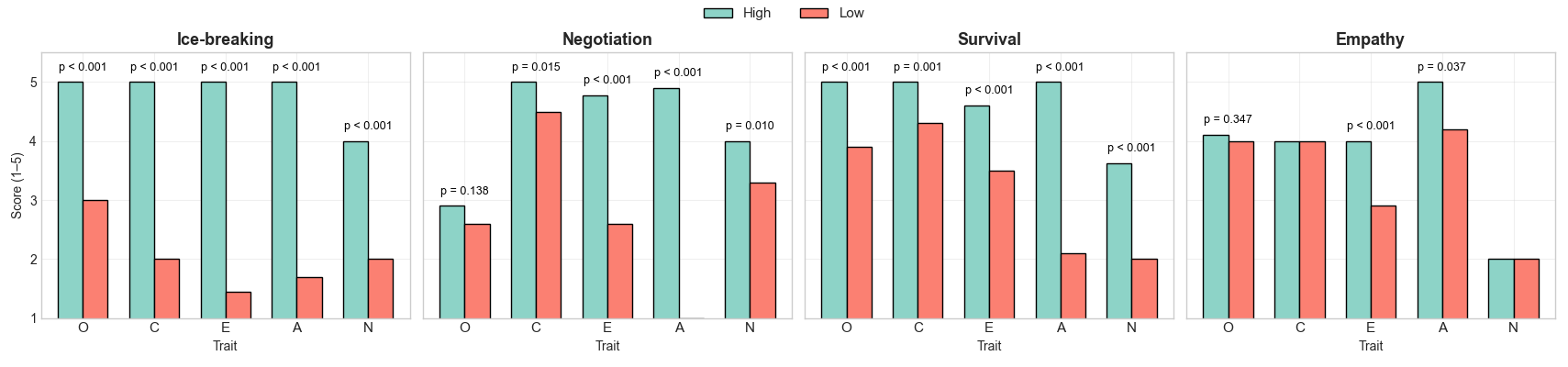}
    \caption{LLM-based Personality Evaluation Result}
    \label{fig:LLM-EVAL}
\end{figure*}

Figure~\ref{fig:LLM-EVAL} shows the LLM-based personality evaluation scores for High and Low conditions across the four dialogue contexts. 
Overall, the results closely aligned with the intended personality manipulation, showing significant differences ($p < .001$) across most traits. 
Agents in the High condition consistently received higher scores on their corresponding traits, indicating that the LLM evaluator effectively recognized the designed personality patterns through linguistic behavior in dialogue.
Across contexts, personality differences were most pronounced in the Ice-breaking and Survival tasks, where the cooperative and emotionally expressive nature of the interaction facilitated clearer trait expression. 
In contrast, Negotiation and Empathy tasks showed weaker or only partial differentiation, likely due to contextual factors such as emotional sensitivity.


\subsection{Emotional Tone Across Dialogue Contexts}

\begin{figure}[h]
    \centering
    \includegraphics[width=1\linewidth]{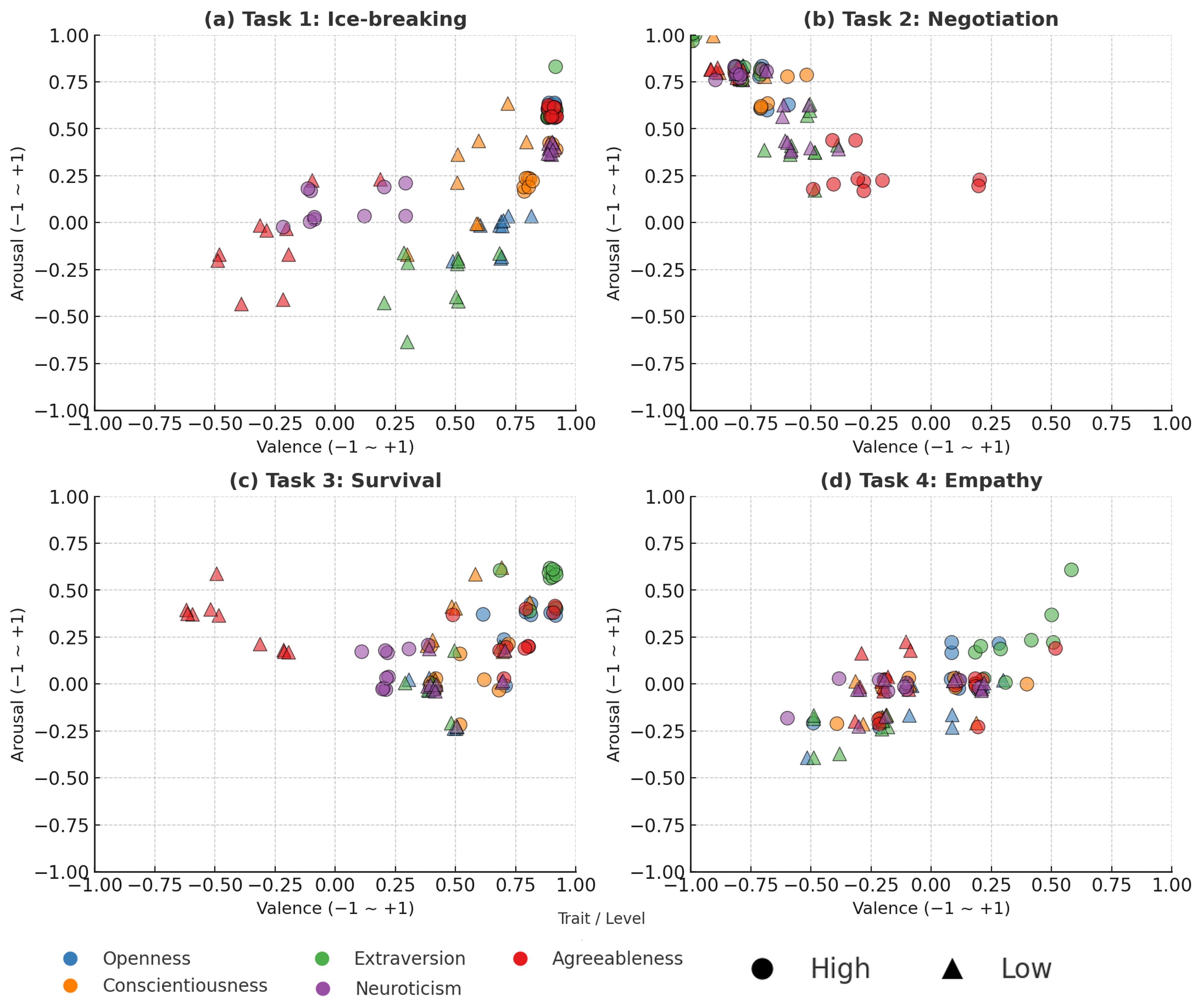}
    \caption{Distribution of Valence–Arousal across four dialogue contexts}
    \label{fig:Emotion-VA-Space}
\end{figure}

To examine affective variation across tasks, 
we compared the Valence–Arousal distribution extracted from each dialogue (Fig.~\ref{fig:Emotion-VA-Space}).  
Distinct patterns were observed across the four contexts.  
The Ice-breaking phase showed generally positive valence and high arousal, reflecting lively and engaging exchanges.  
Negotiation displayed negative valence and moderate arousal, indicating tension and goal conflict.  
The Survival task presented mixed emotions with moderate valence and arousal, consistent with both cooperative and competitive dynamics.  
Finally, the Empathy context showed positive valence and low arousal, suggesting calm and emotionally supportive interactions.

\subsection{Summary: From Linguistic to Behavioral and Emotional Expression (RQ1)}

\begin{figure}[h]
    \centering
    \includegraphics[width=1\linewidth]{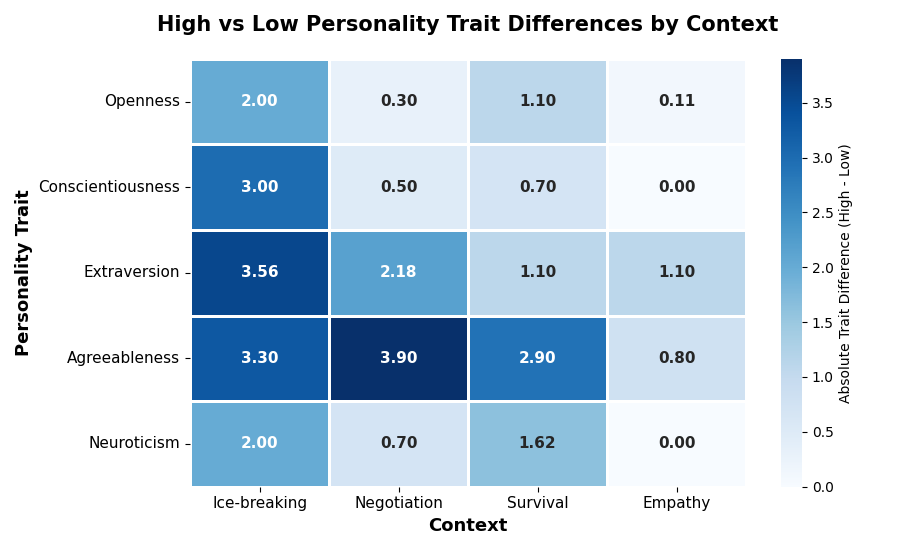}
    \caption{LLM-based personality difference across dialogue contexts. 
Higher values represent stronger trait differentiation (High vs. Low).}
    \label{fig:personality_context}
\end{figure}

Across analyses, personality-related differences emerged consistently but with varying degrees of comparability and expressiveness.  
The LIWC-based linguistic analysis revealed within-trait differences (High vs. Low) but could not be directly compared across traits, 
as each dimension relied on distinct linguistic features.  
In contrast, both the Pre-trained and LLM-based evaluations operated within shared representational spaces, 
allowing for cross-trait and cross-context comparison.  
While the Pre-trained model captured conceptual differentiation among personality dimensions, 
the LLM-based evaluation extended this analysis to behavioral outcomes, visualizing how personality expression varied across dialogue contexts (Fig.~\ref{fig:personality_context}).  
Specifically, trait differentiation was strongest in cooperative settings such as Ice-breaking and Survival, 
and diminished under competitive or emotionally sensitive contexts like Negotiation and Empathy.

Finally, the Valence–Arousal analysis complemented these findings by illustrating the affective tone underlying these interactions.  
Contexts with greater personality differentiation also showed broader emotional variance and higher mean Valence, 
suggesting that positive and engaged affect co-occurred with clearer personality expression.  
These results indicate that personality in LLM-based agents manifests across linguistic, representational, behavioral, and affective levels, 
with context modulating the strength of expression across all dimensions.

\section{Result: Behavioral Expression of Personality (RQ2)}

This section addresses RQ2 — Do personality-driven tendencies extend beyond linguistic style to task-oriented behaviors such as concession-making and agreement outcomes?  
We examine whether personality-driven differences observed in dialogue expression also influence how agents cooperate and adjust their decisions in task-oriented interactions.  
To this end, we analyze two behavioral dimensions: Agreement, indicating whether agents reached a mutual consensus, and Concession, reflecting the degree to which each agent modified its decisions across interaction rounds.  
Concession is further analyzed in two contexts: Negotiation (Refund Offer) and Survival (Sum of Rank Differences; SRD).

\subsection{Agreement Rate (\%)}
As shown in Table~\ref{tab:agreement-table}, the overall agreement rate in the Negotiation task was relatively low,  
reflecting its inherently conflict-driven and competitive nature.  
Because this task required resolving disputes rather than building consensus, agents often failed to reach mutual agreement.  
However, agents high in Agreeableness achieved a 20\% agreement rate,  
higher than those low in Extraversion or low in Neuroticism (each around 10\%).  

In contrast, the Survival task (Table~\ref{tab:agreement-table}) presented a highly collaborative environment with generally higher agreement rates.  
Specifically, agents high in Agreeableness reached consensus in 90\% of cases, compared with 70\% for their low counterparts,  
and those high in Extraversion achieved 80\% agreement versus 40\% for low Extraversion.  

\begin{table*}[h!]
\centering
\begin{subtable}[b]{0.48\linewidth}
    \centering
    \caption{Agreement Rate (\%) in Negotiation Task}
    \begin{tabular}{lcc}
        \toprule
        Personality Dimension & High (\%) & Low (\%) \\
        \midrule
        Openness           & 0.0  & 0.0 \\
        Conscientiousness  & 10.0 & 0.0 \\
        Extraversion       & 10.0 & 20.0 \\
        Agreeableness      & 20.0 & 0.0 \\
        Neuroticism        & 0.0  & 10.0 \\
        \bottomrule
    \end{tabular}
\end{subtable}
\hfill
\begin{subtable}[b]{0.48\linewidth}
    \centering
    \caption{Agreement Rate (\%) in Survival Task}
    \begin{tabular}{lcc}
        \toprule
        Personality Dimension & High (\%) & Low (\%) \\
        \midrule
        Openness           & 90.0 & 50.0 \\
        Conscientiousness  & 50.0 & 60.0 \\
        Extraversion       & 80.0 & 40.0 \\
        Agreeableness      & 90.0 & 70.0 \\
        Neuroticism        & 90.0 & 30.0 \\
        \bottomrule
    \end{tabular}
\end{subtable}
\caption{Agreement rates by personality dimension across two tasks.}
\label{tab:agreement-table}
\end{table*}

\subsection{Concession}

\subsubsection{Negotiation (Refund Offer)}

\begin{figure}[h]
    \centering
    \begin{subfigure}[b]{0.9\linewidth}
        \centering
        \includegraphics[width=\linewidth]{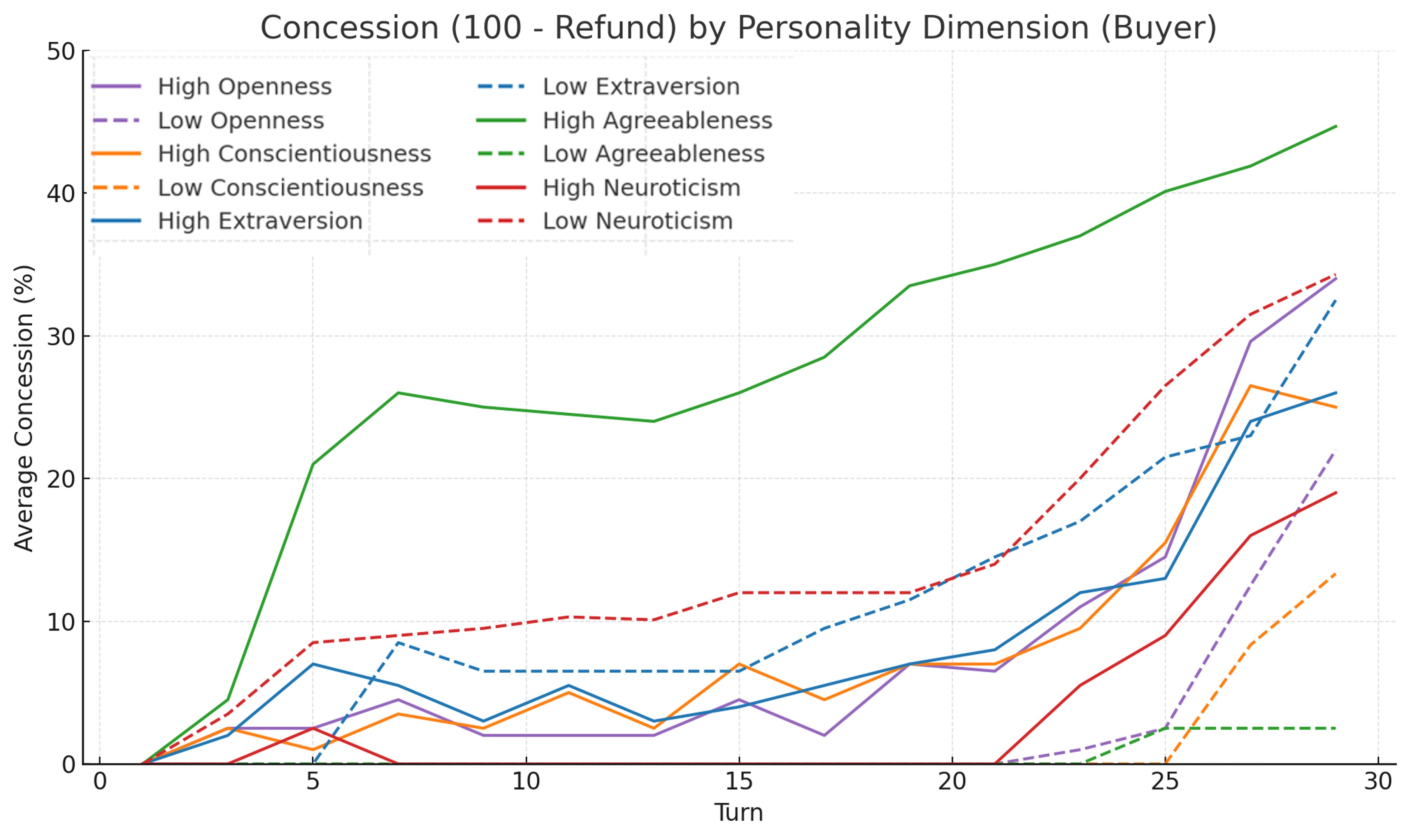}
        \caption{Refund offer trends by buyer personality across negotiation turns.}
        \label{fig:offer_trend}
    \end{subfigure}

    \vspace{0.5em} 

    \begin{subfigure}[b]{0.9\linewidth}
        \centering
        \includegraphics[width=\linewidth]{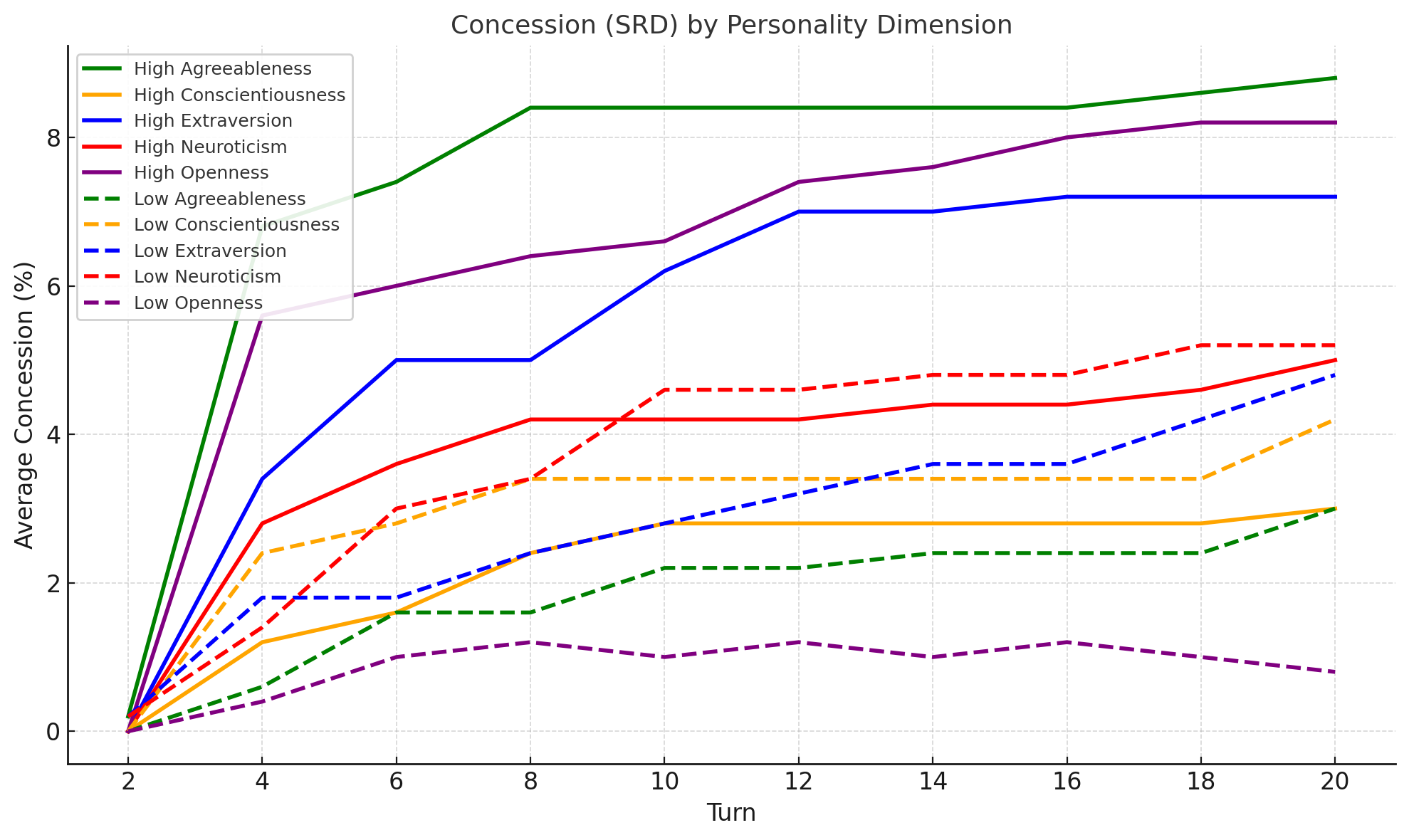}
        \caption{SRD variation by personality dimension across dialogue turns.}
        \label{fig:srd_change}
    \end{subfigure}

    \caption{Behavioral outcomes by personality: (a) refund-based concession in negotiation and (b) SRD-based adaptation in survival.}
    \label{fig:behavior_nego_combined}
\end{figure}

In the negotiation task, concession behavior was evaluated based on the amount of reduction from the initial refund proposal, calculated as (100-Refund). 
As shown in Figure~\ref{fig:offer_trend}, agents high in Agreeableness exhibited the strongest concession pattern, gradually increasing their concession to over 40\% by the final turns.  
In contrast, agents low in Agreeableness and those high in Neuroticism showed minimal change, maintaining concession levels below 10\% for most of the dialogue.  
Openness and Extraversion produced moderate concession curves, converging around 20–25\% toward the end.  
Overall, the results indicate that agreeable and emotionally stable agents are more willing to compromise and adjust their positions during negotiation,  
whereas antagonistic or highly neurotic agents remain more rigid and less responsive to their partner’s demands.  
This behavioral tendency is consistent with the agreement outcomes reported earlier (Table~\ref{tab:agreement-table}),  
where high-Agreeableness agents also achieved higher rates of successful resolution.

\subsubsection{Survival (Sum of Rank Differences; SRD)}

In the survival task, concession was measured by the Sum of Rank Differences (SRD) across decision rounds.  
As shown in Figure~\ref{fig:srd_change}, agents high in Agreeableness and Openness demonstrated steadily increasing SRD values,  
reaching approximately 6 to 7 points in later rounds—more than double that of their low-trait counterparts (below 3).  
Extraverted agents exhibited moderate adaptation, whereas those high in Neuroticism showed unstable, fluctuating changes toward the end of interaction.  
These results suggest that open and affiliative personalities exhibit greater flexibility and adjust their decisions more dynamically in cooperative contexts.

\section{Discussion}

This study examined how personality expression in LLM-based agents extend beyond linguistic patterns to social behaviors and emotional responses.
At the linguistic level, LIWC analysis captured within-trait differences (High vs. Low) but showed limited cross-trait or cross-context variation.
This constraint likely arises because LIWC relies on distinct feature sets for each trait, making direct comparison across dimensions difficult.
In contrast, the Pre-trained and LLM-based evaluations operated within shared representational spaces, allowing for cross-trait and context-sensitive interpretation.
The results revealed that cooperative contexts (Ice-breaking and Survival) amplified Extraversion and Agreeableness,
whereas competitive contexts (Negotiation) heightened Neurotic tendencies and produced more tense, conflict-oriented interactions.
In the Empathy task, emotion regulation was emphasized, leading to more stable and subdued linguistic tone overall.

These findings align with the principles of Whole Trait Theory, which posits that personality is generally stable but dynamically activated depending on social goals and situational cues.
Similarly, the personality expressions observed in LLMs were not fixed reproductions of prompted traits but adaptive, state-level adjustments shaped by task demands and emotional context.
In other words, personality in LLMs emerged not as a static textual artifact but as a socially adaptive response shaped by interactional context.
Finally, emotional analysis provided quantitative support for this pattern.
Agents high in Extraversion and Agreeableness exhibited higher Valence and Arousal, consistent with their cooperative linguistic and behavioral tendencies,
while agents high in Neuroticism displayed lower Valence, reflecting tension and withdrawal in competitive settings.
Together, these results indicate that linguistic, behavioral, and emotional expressions operate in a coherent direction—
showing that personality in LLMs is not merely a scripted construct but contextually modulated and affectively grounded.

\section{Conclusion}

This study examined how personality-conditioned LLM agents adapt their expressive behaviors across conversational contexts.  
Even when given identical personality prompts, their linguistic and behavioral patterns varied systematically depending on the social goals of each task.  
Emotional analyses further revealed that these shifts were accompanied by consistent changes in affective tone, suggesting adaptive alignment between personality expression and contextual demands.  
These findings address an important question for LLM-based dialogue agents—whether such variability reflects inconsistency or context-sensitive adaptation akin to human behavior.  
Our results favor the latter interpretation: the observed variations were not random fluctuations but coherent adjustments to interactional goals and affective conditions.  
However, further work is needed to determine whether these context-sensitive changes are functionally adaptive in the same way that human personality operates.  

While the inclusion of a generic agent provided a baseline for comparison, its presence might still have influenced interactional dynamics.  
Future work will explore interactions between agents with differing personalities to examine personality–personality dynamics and directly test whether LLMs internalize the context-dependent activation mechanism proposed by Whole Trait Theory.  
We also plan to extend this framework to a broader range of social settings and compare LLM-generated behaviors with human conversational data to improve real-world validity.



\bibliography{sample-1col}

\appendix

\section{Personality Prompt}

\subsection{Personality Modulation Prompt}
\begin{tcolorbox}[title=LLM Prompt Snippet]  

\textbf{Extraversion}  
- High Extraversion: outgoing, sociable, energetic, talkative, assertive, enthusiastic  
- Low Extraversion: reserved, quiet, solitary, passive, withdrawn, subdued  

\medskip
\textbf{Agreeableness}  
- High Agreeableness: kind, cooperative, compassionate, warm, trusting, empathetic  
- Low Agreeableness: critical, argumentative, harsh, cold, suspicious, hostile  

\medskip
\textbf{Conscientiousness}  
- High Conscientiousness: organized, reliable, disciplined, responsible, efficient, thorough  
- Low Conscientiousness: careless, disorganized, negligent, lazy, unreliable  

\medskip
\textbf{Neuroticism}  
- High Neuroticism: anxious, moody, insecure, self-conscious, vulnerable, easily stressed  
- Low Neuroticism : calm, relaxed, resilient, confident, secure, steady  

\medskip
\textbf{Openness}  
- High Openness: creative, curious, imaginative, intellectual, adventurous, open-minded  
- Low Openness: rigid, practical, narrow-minded, unimaginative, routine-oriented  
\end{tcolorbox}

\begin{center}
LLM Prompt Snippet with Big Five Personality
\end{center}

\subsection{Personality Evaluation Prompt}
\begin{tcolorbox}[title=LLM Prompt Snippet (Personality Evaluation – Negotiation Context)]

You are a psychologist analyzing personality traits based on the speaker’s
linguistic and behavioral cues observed in the conversation.

\medskip
\textbf{Context:}  
The conversation is between a buyer and a seller addressing a
misunderstanding-related conflict. The speaker seeks a refund and resolution
of the issue.

\medskip
Evaluate the speaker’s personality according to the Big Five dimensions (1–5 scale):

\textbf{Extraversion} — 1 = reserved, quiet \quad | \quad 5 = sociable, talkative, assertive  

\textbf{Agreeableness} — 1 = harsh, critical \quad | \quad 5 = kind, cooperative, empathetic  

\textbf{Conscientiousness} — 1 = careless \quad | \quad 5 = disciplined, reliable, efficient  

\textbf{Neuroticism} — 1 = calm, relaxed \quad | \quad 5 = anxious, moody, insecure  

\textbf{Openness} — 1 = rigid, unimaginative \quad | \quad 5 = creative, curious, open-minded  

\medskip
Analyze the following utterances and infer the speaker’s personality levels.
\end{tcolorbox}

\begin{center}
LLM Prompt Snippet with Negotiation Context
\end{center}

\end{document}